\documentclass[twoside,11pt]{article}

\usepackage{apacite}

\usepackage{jair}

\ShortHeadings{On the evaluation of (meta-)solver approaches}
{Amadini, Gabbrielli, Liu \&  Mauro}
\firstpageno{1}

\usepackage{subcaption}
\usepackage{
beramono 
} 

\usepackage{lineno}
\usepackage[T1]{fontenc}
\usepackage[utf8]{inputenc}
\usepackage{algpseudocode}
\usepackage{listings}
\usepackage{color}
\usepackage{xspace}
\usepackage{url}
\usepackage{multicol}
\usepackage{booktabs}
\usepackage{graphicx}
\usepackage{enumerate}
\usepackage{xcolor}
\usepackage{makecell}
\usepackage{soul}
\usepackage{todonotes}
\usepackage{amsthm}

\usepackage{pgfplotstable}
\usepackage{booktabs}
\usepackage{colortbl}
\usepackage{pgfplots}
\usepackage{pgfplotstable}
\usepackage{tikz}
\usepackage{collcell}
\usepackage{graphicx}
\usepackage{booktabs}
\usepackage{xcolor}

\pgfplotstableset{
    /color cells/min/.initial=0,
    /color cells/max/.initial=1000,
    /color cells/textcolor/.initial=,
    color cells/.code={%
        \pgfqkeys{/color cells}{#1}%
        \pgfkeysalso{%
            postproc cell content/.code={%
                \begingroup
                \pgfkeysgetvalue{/pgfplots/table/@preprocessed cell content}\value
                \ifx\value\empty
                    \endgroup
                \else
                \pgfmathfloatparsenumber{\value}%
                \pgfmathfloattofixed{\pgfmathresult}%
                \let\value=\pgfmathresult
                \pgfplotscolormapaccess
                    [\pgfkeysvalueof{/color cells/min}:\pgfkeysvalueof{/color cells/max}]
                    {\value}
                    {\pgfkeysvalueof{/pgfplots/colormap name}}%
                \pgfkeysgetvalue{/pgfplots/table/@cell content}\typesetvalue
                \pgfkeysgetvalue{/color cells/textcolor}\textcolorvalue
                \toks0=\expandafter{\typesetvalue}%
                \xdef\temp{%
                    \noexpand\pgfkeysalso{%
                        @cell content={%
                            \noexpand\cellcolor[rgb]{\pgfmathresult}%
                            \noexpand\definecolor{mapped color}{rgb}{\pgfmathresult}%
                            \ifx\textcolorvalue\empty
                            \else
                                \noexpand\color{\textcolorvalue}%
                            \fi
                            \the\toks0 %
                        }%
                    }%
                }%
                \endgroup
                \temp
                \fi
            }%
        }%
    }
}

\newtheorem{definition}{Definition}

\usepackage{amsmath,amssymb} 

\usepackage{multirow}

\usepackage{xspace}
\usepackage{booktabs}
\usepackage{tabularx}

\usepackage{csvsimple}
\makeatletter
\csvset{
  autotabularcenter/.style={
    file=#1,
    after head=\csv@pretable\begin{tabular}{|*{\csv@columncount}{c|}}\csv@tablehead,
    table head=\hline\csvlinetotablerow\\\hline,
    late after line=\\,
    table foot=\\\hline,
    late after last line=\csv@tablefoot\end{tabular}\csv@posttable,
    command=\csvlinetotablerow},
  autobooktabularcenter/.style={
    file=#1,
    after head=\csv@pretable\begin{tabular}{*{\csv@columncount}{c}}\csv@tablehead,
    table head=\toprule\csvlinetotablerow\\\midrule,
    late after line=\\,
    table foot=\\\bottomrule,
    late after last line=\csv@tablefoot\end{tabular}\csv@posttable,
    command=\csvlinetotablerow},
}
\makeatother

\newcommand{\bet}{\mathtt{better}}
\newcommand{\unk}{\mathtt{unknown}}

\newcommand{\rtime}{\mathtt{time}}
\newcommand{\val}{\mathtt{obj}}

\newcommand{\sbs}{\ensuremath{\mathit{SBS}}\xspace}
\newcommand{\vbs}{\ensuremath{\mathit{VBS}}\xspace}

\newcommand{\parx}[1]{\mathsf{PAR}_{#1}}

\newcommand{\parten}{\mathsf{\parx{10}}}

\newcommand{\I}{\mathcal{I}}

\renewcommand{\S}{\mathcal{S}}

\usepackage[figuresright]{rotating}

\begin{document}

\title{On the evaluation of (meta-)solver approaches}

\author{
	   \name Roberto Amadini \email roberto.amadini@unibo.it
	   \AND
       \name Maurizio Gabbrielli \email maurizio.gabbrielli@unibo.it \\
       \addr Department of Computer Science and Engineering, \\
       University of Bologna, 
       Italy
       \AND       
       \name Tong Liu \email lteu@icloud.com  \\
       \addr Meituan, \\
       Beijing,
       China
       \AND
       \name Jacopo Mauro \email mauro@imada.sdu.dk \\
       \addr Department of Mathematics and Computer Science,\\
      University of Southern Denmark, Denmark
        }

\researchnote

\maketitle

\begin{abstract}
Meta-solver approaches 
exploits a number of individual solvers to potentially build a better solver. To assess the performance of meta-solvers, one can simply adopt the metrics typically used for individual solvers (e.g., runtime or solution quality), or employ more specific evaluation metrics (e.g., by measuring how close the meta-solver gets to its virtual best performance).
In this paper,
based on some recently published works,
we provide an overview of different performance metrics for evaluating (meta-)solvers, by underlying their strengths and weaknesses.
\end{abstract}

\section{Introduction}

A famous quote attributed to Aristotle says that ``\textit{the whole is greater than the sum of its parts}''. This principle has been applied 
in several contexts, including the field of constraint solving and optimization.
Combinatiorial problems arising from application domains such as
scheduling, manufacturing, routing or logistics can be tackled by combining and leveraging the complementary strengths of different solvers to create a better global \emph{meta-solver}.\footnote{Meta-solvers are sometimes referred in the literature as \emph{portfolio solvers}, because they take advantage of a ``portfolio'' of different solvers.}

Several approaches for combining solvers and hence creating effective meta-solvers have been developed.
Over the last decades we witnessed the creation  
of new Algorithm Selection~\cite{kotthoff2016algorithm} and  Configuration~\cite{DBLP:books/daglib/p/Hoos12} approaches\footnote{A fine-tuned solver can be seen as a meta-solver where we consider different configurations of the same solver as different solvers.} 
that reached peak results in various solving competitions~\cite{satcomp,DBLP:journals/constraints/StuckeyBF10,plancomp}.
To compare different meta-solvers, new competitions were created, e.g., the 2015 ICON  challenge~\cite{kotthoff2015icon} and 2017 OASC competition on algorithm selection~\cite{lindauer2019algorithm}.
However, the discussion of why a particular evaluation metric has been chosen to rank the solvers is lacking.

We believe that further study on this issue is necessary because often meta-solvers are 
evaluated on heterogeneous scenarios, characterized by a different number of problems, 
different timeouts, and different individual solvers from which the meta-solvers approaches are built.
In this paper, starting from some surprising results presented by \citeA{sunnyas2} showing dramatic ranking changes with different, but reasonable, metrics we would like to draw more attention on the evaluation of meta-solvers approaches
by shedding some light on the strengths and weaknesses of different metrics.

\section{Evaluation metrics}

Before talking about the evaluation metrics, we should spend some words on what we need to evaluate: the solvers. In our context, a solver is a program that takes as input the description of a computational problem in a given language, and returns an observable \textit{outcome} providing zero or more solutions for the given problem. For example, for decision problems the outcome may be simply ``yes'' or ``no'' while for optimization problems we might be interested in the sub-optimal solutions found along the search.
An \emph{evaluation metric}, or performance metric, is a function mapping the outcome of a solver on a given instance to a number representing ``how good'' the solver on this instance is.

An evaluation metric is often not just defined by the output of the (meta-)solver but can also be influenced by other actors such as the computational resources available, the problems on which we evaluate the solver, and the other solvers involved in the evaluation. For example, 
it is often unavoidable to set a \emph{timeout} $\tau$ 
on the solver's execution when there is no guarantee of termination in a reasonable amount of time (e.g., NP-hard problems).
Timeouts makes the evaluation feasible, but inevitably couple the evaluation metric to the execution context.
For this reason, the evaluation of a meta-solver should also take into account the  \emph{scenario} that encompasses the solvers to evaluate, the instances used for the validation, and the timeout. 
Formally, at least for the purposes of this paper, we can define a scenario  as a triple
$(\I,\S,\tau)$ where: $\I$ is a set of problem instances, $\S$ is a set of \textit{individual} solvers, $\tau \in (0,+\infty)$ is a timeout such that the outcome of solver $s \in S$ over instance $i \in \I$ is always measured in the time interval $[0,\tau)$.

Evaluating meta-solvers over heterogeneous scenarios is complicated by the fact that the set of instances, solvers and the timeout can have high variability. 
As we shall see in Sect.~\ref{sec:opt}, things are even trickier in scenarios including optimization problems. 


\subsection{Absolute vs relative metrics}

A sharp distinction 
between evaluation metrics can be drawn depending on whether their value depend on the outcome of other solvers or not. We say that an evaluation metric is \emph{relative} in the former case, \emph{absolute} otherwise.
For example, a well-known absolute metric is the
\emph{penalized average runtime} with penalty $\lambda \geq 1$ (PAR$_{\lambda}$) that compares the solvers by using the average solving runtime and penalizes the timeouts with $\lambda$ times the timeout.

Formally, let $\rtime(i,s,\tau)$ be the function returning the runtime of solver $s$ on instance $i$ with timeout $\tau$, assuming $\rtime(i,s,\tau)=\tau$ if $s$ cannot solve $i$ before the timeout $\tau$. 
For optimization problems, we consider the runtime as the time taken by $s$ to solve $i$ to optimality\footnote{If $s$ cannot solve $i$ to optimality before $\tau$, then $\rtime(i,s)=\tau$ even if sub-optimal solutions are found.} assuming w.l.o.g.~that an optimization problem is always 
a minimization problem. We can define PAR$_{\lambda}$ as follows.
\begin{definition}[Penalized Average Runtime]\label{def:par}
Let $(\I,\S,\tau)$ be a scenario, the PAR$_{\lambda}$ score of solver $s \in \S$ over $\I$ 
is given by $\dfrac{1}{|\I|}\sum\limits_{i \in \I} \mathtt{par}_{
	\lambda}(i, s, \tau)$ where:
\[
\mathtt{par}_{\lambda}(i, s, \tau) = \begin{cases}
\rtime(i, s, \tau) & \text{if $\rtime(i, s, \tau) < \tau$}\\
\lambda \cdot \tau & \text{otherwise.}
\end{cases}
\]
\end{definition}

Well-known PAR measures are, e.g., the $\parx{2}$ adopted in the SAT competitions \cite{satcomp}
or the $\parx{10}$ used by \citeA{lindauer2019algorithm}.
Evaluating (meta-)solvers with scenarios having different timeouts should imply a \emph{normalization} of PAR$_{\lambda}$ in a fixed range to avoid misleading comparisons.

Another absolute metric for decision problems is the number (or percentage) of instances solved where ties are broken by favoring the solver minimizing the average running time, i.e., minimizing the $\parx{1}$ score. This metric has been used in various tracks of the planning competition \cite{plancomp}, the XCSP competition \cite{XCSPcomp}, and the QBF evaluations \cite{QBFEVAL}.



A well-established relative metric is instead the \emph{Borda count}, adopted for example by the MiniZinc Challenge~\cite{mzn-challenge} for both single solvers and meta-solvers. 
The Borda count is a family of voting rules that can be applied to the evaluation of a solver by considering the comparison as an election where the solvers are the candidates and the problem instances are the voters.
The MiniZinc challenge uses a variant of Borda\footnote{In the original definition, the lowest-ranked candidate gets 0 points, the next-lowest 1 point, and so on.} where each solver scores points proportionally to the number of solvers it beats.
Assuming that $\val(i,s,t)$ is the best objective value found by solver $s$ on optimization problem $i$ at time $t$, with $\val(i,s,t)=\infty$ when no solution is found at time $t$, the MiniZinc challenge score is defined as follows.
\begin{definition}[MiniZinc challenge score]\label{def:mznc}
Let $(\S,\I,\tau)$ be a scenario where $\I=\I_{dec} \cup \I_{opt}$ with $\I_{dec}$ decision problems and $\I_{opt}$ optimization problems.
The MiniZinc challenge (MZNC) score of 
$s \in \S$ over $\I$ is $\sum\limits_{i \in \I, s'\in S\setminus\{s\}} m_s(i,s',\tau)$ where:
\[
m_s(i, s',\tau) = \begin{cases}
0 & \text{if $\unk(i,s) \vee \bet(i,s',s)$}\\
1 & \text{if $\bet(i,s,s')$}\\
0.5 & \text{if $\rtime(i,s,\tau)=\rtime(i,s',\tau)$} \\
~ &   \text{and $\val(i,s,\tau)=\val(i,s',\tau)$}\\
\dfrac{\rtime(i,s',\tau)}{\rtime(i,s,\tau)+\rtime(i,s',\tau)} & \text{otherwise}
\end{cases}
\]
where the predicate $\unk(i,s)$ holds if $s$ does not produce a solution within the timeout:
$$\unk(i,s) = (i \in \I_{dec} \wedge \rtime(i,s,\tau)=\tau) \vee (i \in \I_{opt} \wedge \val(i,s,\tau)=\infty)$$
and  $\bet(i,s,s')$ holds if $s$ finishes earlier than $s'$ or it produces a better solution:
$$\bet(i,s,s') = \rtime(i,s,\tau) < \rtime(i,s',\tau) = \tau \vee \val(i,s,\tau) < \val(i,s',\tau)$$
\end{definition}
This is clearly a relative metric because changing the set of available solvers can affect the MiniZinc scores.

To handle the disparate nature of the scenarios when comparing meta-solvers approaches, the evaluation function adopted in the ICON and OASC challenges was relative: 
the \emph{closed gap score}. This metric assigns to a meta-solver a value in $[0,1]$ proportional to how much it closes the gap between the best individual solver available, or \textit{single best solver} (\sbs), and the \textit{virtual best solver} (\vbs), i.e., an oracle-like meta-solver always selecting the best individual solver. The closed gap is actually a ``meta-metric'', defined in terms of another evaluation metric.
Formally, if $(\I,\S,\tau)$ is a scenario and $m$ an evaluation metric to minimize, we have $m(i,\vbs,\tau)=\min\{m(i,s,\tau) \mid s \in \S\}$ for each $i \in \I$ 
and $\sbs = \underset{s\in\S}{\operatorname{argmin}} \sum_{i\in\I}\,m(i,s,\tau)$.
We can define the closed gap as follows.

\begin{definition}[Closed gap]\label{def:cgap}
Let $m$ be an evaluation metric to minimize for a scenario $(\I,\S,\tau)$, and let
$m_\vbs = \sum_{i\in\I} m(i,\vbs,\tau)$ and $m_\sbs = \sum_{i\in\I} m(i,\sbs,\tau)$.
The \emph{closed gap} of a (meta-)solver $s$ w.r.t.~$m$ on that scenario is:
\[
\dfrac{m_\sbs-\sum_{i\in\I} m(i,s,\tau)}{m_\sbs-m_\vbs}
\]
\end{definition}
If not specified, we will assume the closed gap computed w.r.t.~the $\parten$ score as done in the AS challenges 2015 and 2017.\footnote{In the 2015 edition, the closed gap was computed as $1-\frac{m_\sbs-m(i,s,\tau)}{m_\sbs-m_\vbs} = \frac{m_\vbs-m(i,s,\tau)}{m_\vbs-m_\sbs}$.}

\begin{table}[t]
	\caption{Comparison ASAP vs RF. The MZNC column reports the average MZNC score per scenario. Negative scores are in bold font.}
	\label{tab:asap_rf}
	\centering
\begin{tabular}{l|cc|cc|cc}
~ & \multicolumn{2}{c|}{Closed gap} & \multicolumn{2}{c|}{MZNC} & \multicolumn{2}{c}{Better than other}\\
Scenario & ASAP & RF & ASAP & RF & ASAP & RF \\
\hline
ASP-POTASSCO & 0.7444 & 0.5314 & 2.2235 & 2.6163 & 275 & 671\\ 
BNSL-2016 & 0.8463 & 0.7451 & 1.2830 & 3.0250 & 98 & 993\\ 
CPMP-2015 & 0.6323 & 0.1732 & 2.0501 & 2.3660 & 137 & 334\\ 
CSP-MiniZinc-Time-2016 & 0.6251 & 0.2723 & 2.1552 & 2.7214 & 17 & 53\\ 
GLUHACK-2018 & 0.4663 & 0.4057 & 1.9040 & 2.4528 & 62 & 147\\ 
GRAPHS-2015 & 0.758 & \textbf{-0.6412} & 2.3045 & 3.3731 & 489 & 3663\\ 
MAXSAT-PMS-2016 & 0.5734 & 0.3263 & 1.4747 & 2.8616 & 66 & 439\\ 
MAXSAT-WPMS-2016 & 0.7736 & \textbf{-1.1826} & 1.5168 & 2.4043 & 126 & 386\\ 
MAXSAT19-UCMS & 0.6583 & \textbf{-0.2413} & 2.0893 & 2.5189 & 145 & 269\\ 
MIP-2016 & 0.35 & \textbf{-0.3626} & 2.4035 & 2.4239 & 81 & 105\\ 
QBF-2016 & 0.7568 & \textbf{-0.1366} & 1.8642 & 2.7154 & 193 & 467\\ 
SAT03-16\_INDU & 0.3997 & 0.1503 & 2.1508 & 2.5812 & 491 & 1116\\ 
SAT12-ALL & 0.7617 & 0.6528 & 1.6785 & 2.8250 & 262 & 1227\\ 
SAT18-EXP & 0.5576 & 0.3202 & 1.9239 & 2.4998 & 61 & 164\\ 
TSP-LION2015 & 0.4042 & \textbf{-19.1569} & 2.4352 & 2.6979 & 1115 & 1949\\
\hline
Tot. & 9.3077 & \textbf{-18.1439} & 29.4573 & 40.0826 & 3618 & 11983\\ 
Tot.$-$TSP-LION2015 & 8.9035 & 1.013 & 27.0221 & 37.3846 & 2503 & 10034\\
\end{tabular}
\end{table}

\subsection{A surprising outcome}

An interesting outcome reported in \citeA{sunnyas2} was the profound difference between the closed gap and the MiniZinc challenge scores.
\citeauthor{sunnyas2} compared the performance of six meta-solvers approaches
across 15 decision-problems scenarios taken from ASlib~\cite{bischl2016aslib} and coming from heterogeneous domains such as Answer-Set Programming, Constraint Programming, Quantified Boolean Formula, Boolean Satisfiability.

Tab.~\ref{tab:asap_rf} reports the performance of ASAP and RF, respectively the best approach according to the closed gap score and the MZNC score. 
The scores in the four leftmost columns clearly show a remarkable difference rank if we swap the evaluation metric. With the closed gap, ASAP is the best approach and RF the worst among all the meta-solvers considered, while with the MZNC score RF climbs to the first position while ASAP drops to the last position.


Another thing that catches the eye in Tab.~\ref{tab:asap_rf} is the presence of \emph{negative  scores}. This happens because, by definition, the closed gap has upper bound 1 
(no meta-solver can improve the \vbs) but not a fixed lower bound. Hence, when the performance of the meta-solver is worse than the performance of the single best solver, 
the closed gap drops below zero.
While on a first glance this seems reasonable---meta-solvers should perform no worse than the individual solvers---it is worth noting that the penalty for performing worse than the SBS  also depends on the denominator $m_\sbs-m_\vbs$. This means that in scenarios where the performance of the \sbs is close to the perfect performance of the \vbs this penalty can be significantly magnified.
The TSP-LION2015 scenario is a clear example: the RF approach gets a penalization of more than 19 points, meaning that RF should perform flawlessly in about 20 other scenarios to expiate this punishment.
In fact, in TSP-LION2015 the $\parten$ distributions of $\sbs$ and $\vbs$ are very close:
the \sbs is able to solve 99.65\% of the instances solved by the \vbs, leaving little room for improvement. RF scores -19.1569 while still solving more than 90\% of the instances of the scenario and having a difference with ASAP of slightly more than 5\% instances solved.


Why are the closed gap and the MZNC rankings so different?
Looking at the rightmost two columns in Tab.~\ref{tab:asap_rf} showing, for each scenario, the number of instances where an approach is faster than the other,
one may conclude that RF is far better than ASAP. In all the scenarios the number of instances where its runtime is lower than ASAP runtime is greater than the number of instances where ASAP is faster. Overall, it is quite impressive to see that RF beats ASAP on 11983 instances while ASAP beats RF on 3618 times only.

An initial clue of why this happens is revealed in \citeA{sunnyas2}, where a parametric version of MZNC score is used. In practice, Def.~\ref{def:mznc} is generalized by 
assuming the performance of two solvers equivalent if their runtime difference is below a given time threshold $\delta$.\footnote{Formally, if $|\rtime(i,s,\tau)-\rtime(i,s',\tau)| \leq \delta$ 
then both $s$ and $s'$ scores 0.5 points---note that if $\delta=0$ we get the original MZNC score as in Def.~\ref{def:mznc}.}
This variant was considered because a time difference of few seconds could be considered irrelevant if solving a problem can take minutes or hours.
\begin{figure}[t]
	\includegraphics[width=1\linewidth]{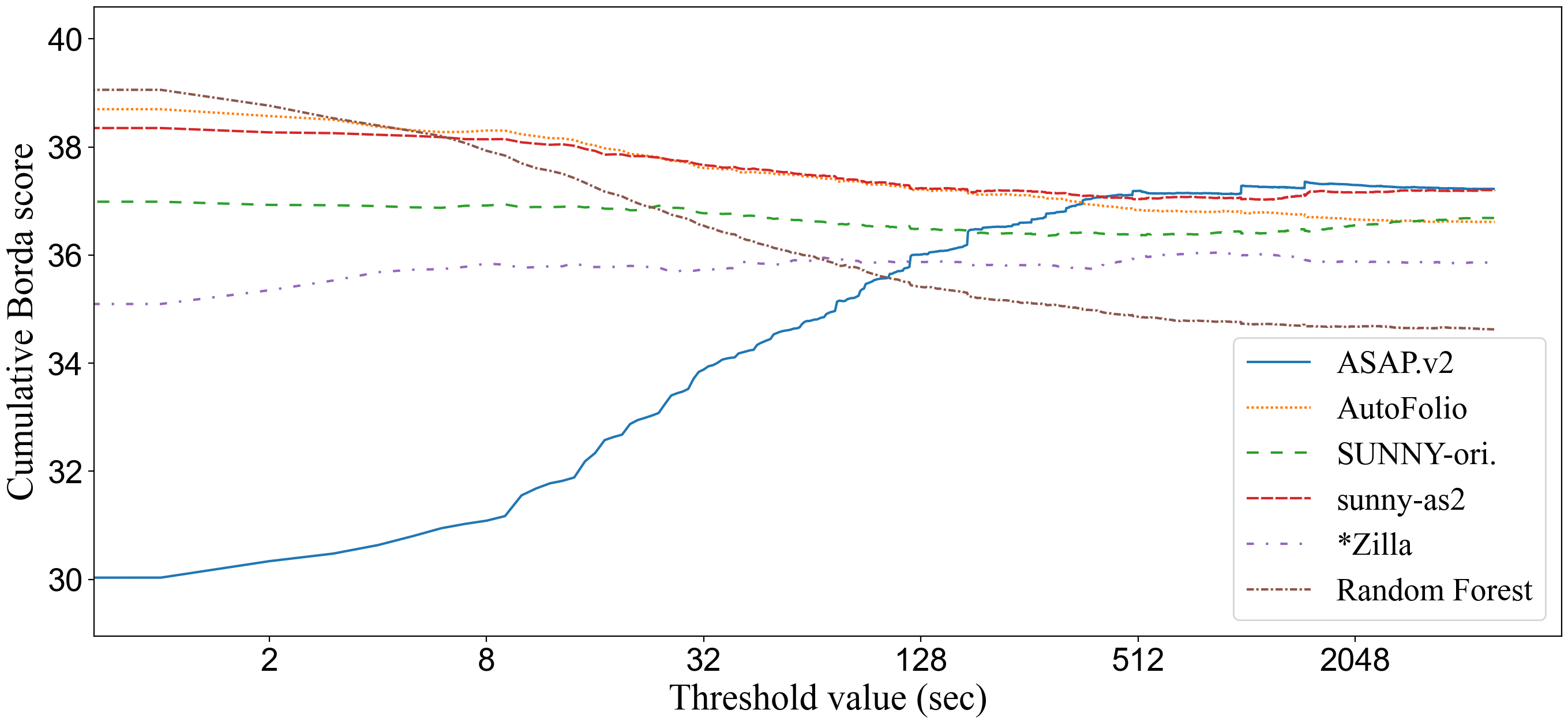}
	\caption{Cumulative Borda count by varying the $\delta$ threshold.\label{fig:borda_jair}}
\end{figure}
The parametric MZNC score is depicted in Fig.~\ref{fig:borda_jair}, where 
different thresholds $\delta$ are considered on the x-axis. It is easy to see how the performance of ASAP and RF reverses when $\delta$ 
increases: ASAP bubbles from the bottom to the top, while RF gradually sinks to the bottom.

Let us further investigate this anomaly.
%
\begin{figure}[t]
	\includegraphics[width=1\linewidth]{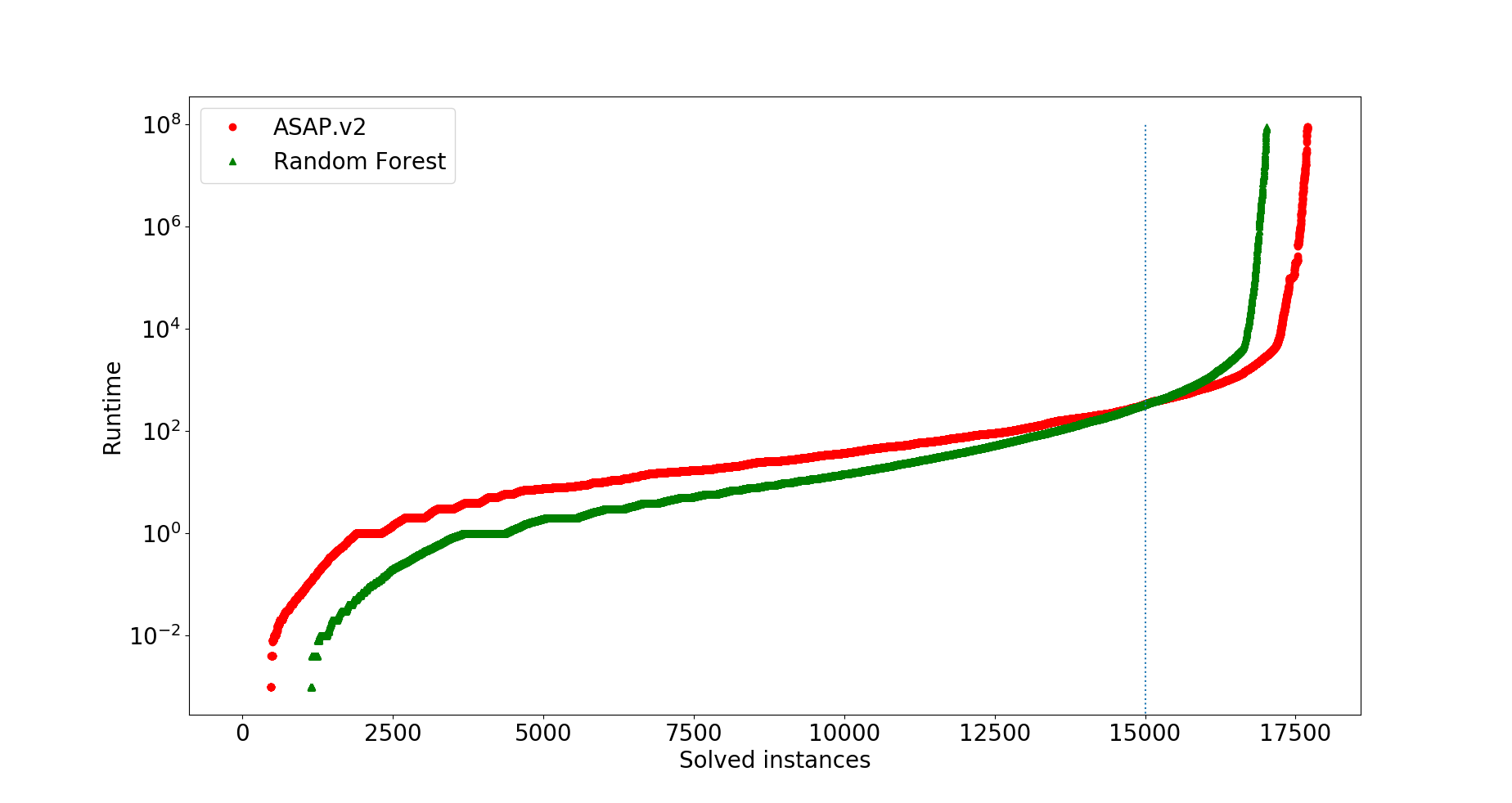}
	\caption{Solve instances difference ASAP vs RF.}\label{fig:asap_rf_solved}
\end{figure}
Fig.~\ref{fig:asap_rf_solved} shows the runtime distributions of the instances solved by ASAP and RF, sorted by ascending runtime. We can see that ASAP solves more instances, but for around 15k instances RF is never slower than ASAP.
Summarizing, ASAP solves more instances but RF is in general quicker when it solves an (often easy) instance. This entails the significant difference between closed gap and Borda metrics.

In our opinion, on the one hand, it is fair to think that ASAP performs better than RF on these scenarios. The MZNC score seems to over-penalize ASAP w.r.t.~RF. Moreover, from Fig.~\ref{fig:borda_jair} 
we can also note that for $\delta \leq 103$ the parametric MZNC score of RF is still better, but 103 seconds looks quite a high threshold to consider two performances as equivalent.
On the other hand, the closed gap score can also be over-penalizing due to  negative outliers.

We also would like to point out that the definitions of \sbs found in the literature do not clarify how it is computed in scenarios where the set of instances $\I$ is split into test and training sets. Should the \sbs be computed on the instances of the training set, the test set, or the whole dataset $\I$? One would be inclined to use the test set to select the \sbs, but this choice might be problematic because the test set is usually quite small w.r.t.~the training set when using, e.g., cross-validation methods. In this case issues with negative outliers might be amplified.
If not clarified, this could lead to confusion. For example, 
in the 2015 ICON challenge the \sbs was computed by considering the entire dataset (training and testing instances together). In the 2017 OASC instead the \sbs was originally computed on the test set of the scenarios, but then the results were amended by computing the \sbs on the training set.

An alternative to the above metrics is the \emph{speedup}
of a single solver w.r.t.~the \sbs or the \vbs, i.e., how much a meta-solver can improve a 
baseline solver.
Tab.~\ref{tab:par_speed} reports, using the data of \cite{sunnyas2}, for each meta-solver $s$ in a scenario $(\I,\S,\tau)$ 
the average speedup computed as $\dfrac{1}{|\I|}\sum_{i\in\I} \frac{\rtime(i,\vbs,\tau)}{\rtime(i,s,\tau)}$.
%
Unlike the closed gap, that has no lower bound, the speedup always falls in $[0,1]$ with bigger values meaning better performance.
We compared this with the \textit{average normalized runtime} score, computed as 
$1-\frac{1}{|\I|}\sum_{i\in\I} \frac{\rtime(i,s,\tau)}{\tau}$ and the average closed gap score w.r.t~$\parx{1}$. We use $\parx{1}$ instead of $\parx{10}$ to be consistent with speedup and normalized runtime, which do not get any penalization.
\begin{table}[t]
	\caption{Average closed gap, speedup, and normalized runtime. Peak performance in bold font.}
	\label{tab:par_speed}
	\centering
	\begin{tabular}{l|c|c|c}
Meta-solver & Closed gap & Speedup & Norm. runtime \\
\hline
ASAP & \textbf{0.4866} & 0.4026 & 0.8829\\ 
sunny-as2 & 0.4717 & \textbf{0.4122} & \textbf{0.8879}\\ 
autofolio & 0.4713 & 0.4110 & 0.8855\\ 
SUNNY-original & 0.4412 & 0.3905 & 0.8790\\ 
*Zilla & 0.3416 & 0.3742 & 0.8753\\ 
Random Forest & -0.1921 & 0.3038 & 0.8507\\ 
	\end{tabular}
\end{table}

The rank with speedup and normalised runtime is the same. The podium changes if we use the closed gap: in this case sunny-as2 and autofolio lose one position while ASAP rises from third to first position. However, as we shall see in the next section, the generalization of these metrics to optimization problems is not trivial.


\subsection{Optimization problems}
\label{sec:opt}
So far we have mainly talked about evaluating meta-solvers on decision problems. While the MZNC score takes into account also optimization problems, for the closed gap, (normalized) runtime, and speedup the generalization is not as obvious as it might seem.
Here using the runtime might not be the right choice: often a solver cannot prove the optimality of a solution, even when it actually finds it. Hence, the obvious alternative 
is to consider just the objective value of a solution. But this value needs to be  \emph{normalized}, and to do so what bounds should we choose?
Furthermore:
how to reward a solver that actually proves the optimality of a solution? And how to penalize solvers that cannot find any solution?

If solving to optimality is not rewarded, metrics such as the
ratio score of the satisfiability track of the planning competition can be used.\footnote{This track includes optimization problems where the goal is to minimize the length of a plan.}
This score is computed as the ratio between the best known solution and the best objective value found by the solver, giving 0 points in case no solution is found.

A different metric that focuses on quickly reaching good solution is the \emph{area} score, introduced in the MZNC starting from 2017. This metric computes the integral of a step function of the solution value over the runtime horizon. Intuitively, a solver that 
finds good solutions earlier can outperform a solver that finds better solutions much later in 
the
solving stage.

Other attempts have been proposed to take into account the objective value and the running times.
For example, the ASP competition \cite{ASPcomp} adopted an elaborate scoring system that combines together the percentage of instances solved within the timeout, the evaluation time, and the quality of a solution.
Similarly, \citeA{paper_amai} proposed a relative metric where each solver $s$ gets a reward in $\{0\} \cup [\alpha,\beta] \cup \{1\}$ according to the objective value $\val(s,i,\tau)$ of the best solution it finds, with $0\leq\alpha\leq\beta\leq1$. If no solution is found then $s$ scores 0, if it solves $i$ to optimality it scores 1, otherwise
the score is computed by linearly scaling $\val(s,i,\tau)$ in $[\alpha,\beta]$ according to the best and worst objective values find by any other available solver on problem $i$.

\subsection{Randomness and aggregation}
We conclude the section with some remarks about randomness and data aggregation.

When evaluating a meta-solver $s$ on scenario $(\I,\S,\tau)$, it is common practice to partition $\I$ into a training set $\I_{tr}$, on which $s$ ``learns'' how to leverage its individual solvers, and a test set $\I_{ts}$ where the performance of $s$ on unforeseen problems is measured. In particular, to prevent overfitting, it is possible to use a $k$-fold cross validation by first splitting 
$\I$ into $k$ disjoint folds, and then using in turn one fold as test set and the union of the other folds as training set. In the AS challenge 2015~\cite{lindauer2019algorithm} the submissions were indeed evaluated with a 10-fold cross validation, while in the OASC in 2017 the dataset of the scenarios was divided only in one test set and one training set.
As also underlined by the organizers of the competition, this is risky 
because 
it may reward a lucky meta-solver performing well on that split but poorly on other splits.

Note that so far we have
always assumed deterministic solvers, i.e., solvers always providing the same outcome if executed on the same instance in the same execution environment. Unfortunately, the scenario may contain randomized solvers potentially producing different results with a high variability. In this case, solvers should be evaluated over a number of runs and particular care must be taken because the assumption that a solver can never outperform the VBS would be no longer true.

A cautious choice to decrease the variance of model predictions 
would be to repeat the $k$-fold cross validation $n > 1$ times with different random splits.
However, this might imply a tremendous computational effort---the training phase of a  meta-solver might take hours or days---and therefore a significant energy consumption.
This issue is becoming an increasing concern. For example, in their recent work \citeA{DBLP:conf/cp/MatriconAFSH21} 
propose an approach to early stop running an individual solver that it is likely to perform worse than another solver on a subset of the instances of the scenario.
In this way, less resources are wasted for solvers that most likely will not bring any 
improvement.

%

Finally, we spend a few words on the aggregation of the results. 
It is quite common to use the arithmetic mean, or just the sum, when it comes to aggregate the outcomes of a meta-solver over different problems of the same scenario (e.g., when evaluating the results on the $n \cdot k$ test sets of a $k$-fold cross validation repeated $n$ times). 
The same applies when evaluating different scenarios.
The choice of how to aggregate the metric values into a unique value should however be motivated since the arithmetic mean can lead to misleading conclusions when summarizing normalized benchmark~\cite{DBLP:journals/cacm/FlemingW86}.
For example, 
to amortize the effect of outliers,
one may use the median or use the geometric mean to average over normalized numbers.



\section{Conclusions}
As it happens in many other fields, the choice of reasonable metrics can have divergent effects on the assessment of (meta-)solvers. While these issues are mitigated when comparing individual solvers in competitions having uniform scenarios in term of size, difficulty, and nature, the comparison of meta-solver approaches poses new challenges due to diversity of the scenarios on which they are evaluated. Although it is impossible to define a \textit{fits-all} metric, we believe that we should aim at more robust metrics avoiding as much as possible the under- and over-penalization of meta-solvers. 

Particular care should be taken when using \emph{relative} measurements, because the risk is to amplify small performance variations  into large differences of the metric's value.
Presenting the results in terms of orthogonal evaluation metrics allows a better understanding of the (meta-)solvers performance, and these insights may help the researchers to build meta-solvers that better fit their needs, as well as to prefer an evaluation metric over another. Moreover, well-established metrics may be combined into hybrid ``meta-metrics'' 
folding together different performance aspects and handling the possible presence of randomness.

\newpage

\newpage

\bibliography{biblio}
\bibliographystyle{apacite}

\end{document}